# A New Method to Extract Dorsal Hand Vein Pattern using Quadratic Inference Function

Maleika Heenaye- Mamode Khan
*Department of Computer Science and Engineering,
University of Mauritius,
Mauritius.*

Naushad Ali Mamode Khan
*Department of Mathematics,
University of Mauritius,
Mauritius .*

*Abstract*—Among all biometric, dorsal hand vein pattern is attracting the attention of researchers, of late. Extensive research is being carried out on various techniques in the hope of finding an efficient one which can be applied on dorsal hand vein pattern to improve its accuracy and matching time. One of the crucial step in biometric is the extraction of features. In this paper, we propose a method based on quadratic inference function to the dorsal hand vein features to extract its features. The biometric system developed was tested on a database of 100 images. The false acceptance rate (FAR), false rejection rate (FRR) and the matching time are being computed.

## I. INTRODUCTION

There is an increasing interest for biometric in the research community since traditional verification methods such as passwords, personal identification numbers (PINS), magnetic swipe cards, keys and smart cards offer very limited security and are unreliable[1],[2]. Biometric which involves the analysis of human biological, physical and behavioral characteristics is being developed to ensure more reliable security. The most popular biometric features that are used are fingerprints, hand geometry, iris scans, faces, as well as handwritten signatures. Recently dorsal hand vein pattern biometric is attracting the attention of researchers and is gaining momentum. Anatomically, aside from surgical intervention, the shape of vascular patterns in the back of the hand is distinct from each other [3], [4], [5]. Veins are found below the skin and cannot be seen with naked eyes. Its uniqueness, stability and immunity to forgery are attracting researchers. These feature makes it a more reliable biometric for personal identification [1]. Furthermore, the state of skin, temperature and humidity has little effect on the vein image, unlike fingerprint and facial feature acquirement [6]. The hand vein biometrics principle is non- invasive in nature where dorsal hand vein pattern are used to verify the identity of individuals [7]. Vein pattern is also stable, that is, the shape of the vein remains unchanged even when human being grows.

Extensive researches are carried out on vein patterns and researchers are striving hard to find methods and techniques to develop dorsal hand vein security system. Any biometric system consists of four main steps namely the preprocessing, feature extraction, processing and matching phase. Feature extraction is a crucial step in biometric system and its capability directly influence the performance of the system.

In this work, the method proposed aims at reducing the dimension of the training set by building an adaptive estimating equation or a quadratic inference function [8],[9] that combines the covariance matrix and the vectors in the training set. The organisation of the paper is as follows: in section II we describe the pre-processing phases applied on the hand dorsal vein pattern, feature extraction using quadratic inference function is presented in section III, we explain the vein pattern matching in section IV and finally the experimental results are presented in chapter V.

## II. PREPROCESSING PHASES APPLIED ON THE DORSAL HAND VEIN BIOMETRIC

First of all, it is necessary to obtain the vein pattern in the image captured. The preprocessing phases which consist of image acquisition, hand segmentation, vein pattern segmentation, noise filtering and thinning of the vein pattern are applied on the hand dorsal vein pattern.

### A. Image Acquisition

Vein pattern is found beneath the skin and is invisible to the naked eye. Up to now, there is currently no publicly available hand vein pattern database available to the research community [2]. Each researcher has to capture their own images by devising their own setup. Vein images can only be captured by using either near infrared or far infrared light. However, according to research better quality images can be obtained using near infrared light [10]. A thermal camera or alternative setup like using a charge coupled device (CCD) with alternative devices can be used to capture the dorsal hand vein pattern.

In this work a database of 100 hand dorsal vein pattern was obtained by a group of researcher, Prof.Ahmed Badawi from University of Tennessee Knoxville [7]. The images were taken with a commercially available conventional charge couple device (CCD) monochrome camera. The hand was presented as a clenched fist with the thumb and all the other fingers hidden. In the setup the intensity of the IR source is attenuated by the use of diffusing paper, which also helps for obtaining an equally distributed illumination on the hand area. A frame grabber is used to capture the image for computer processing. Images were captured using a 320W x 240H pixels video digitizer with a gray- scale resolution of 8- bits per pixel [7], [11], [12]. All the images obtained are of width 320 pixels and of height 240 pixels. It is to be noted that all image templates are of the same size. The figure below shows one of the samples of the dorsal hand vein pattern obtained.

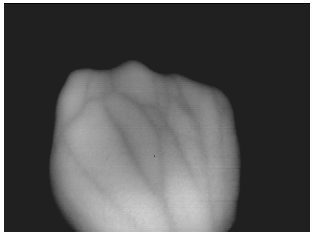

Figure 1. Sample of a hand dorsal vein pattern

### B. Hand and Vein Pattern Segmentation

There is a need to remove all the unnecessary information obtained during data capture. When a hand image is obtained, the hand background is first segmented from the image. For hand segmentation, morphological operations are applied on the hand image to estimate the background of the hand region. The two morphological operations are dilation and erosion, where dilation is an operation that "grows" objects and erosion "thins" objects in a binary image. Erosion followed by dilation was used and this creates an important morphological transformation called opening. The opening of an image X by structuring element B is denoted by XoB and is defined as follows [13]:

$$X \circ B = (X \oplus B) \oplus B \qquad (1)$$

The background was then subtracted from the original image. This allows us to obtain the region of interest. The contrast that varies all over the vein image has been adjusted. After this operation the hand is being segmented.

In order to obtain the vein pattern, the image is then thresholded. Thresholding is the most common segmentation method which is computationally fast and inexpensive. Figure 2 shows the resulting images after applying all the above steps.

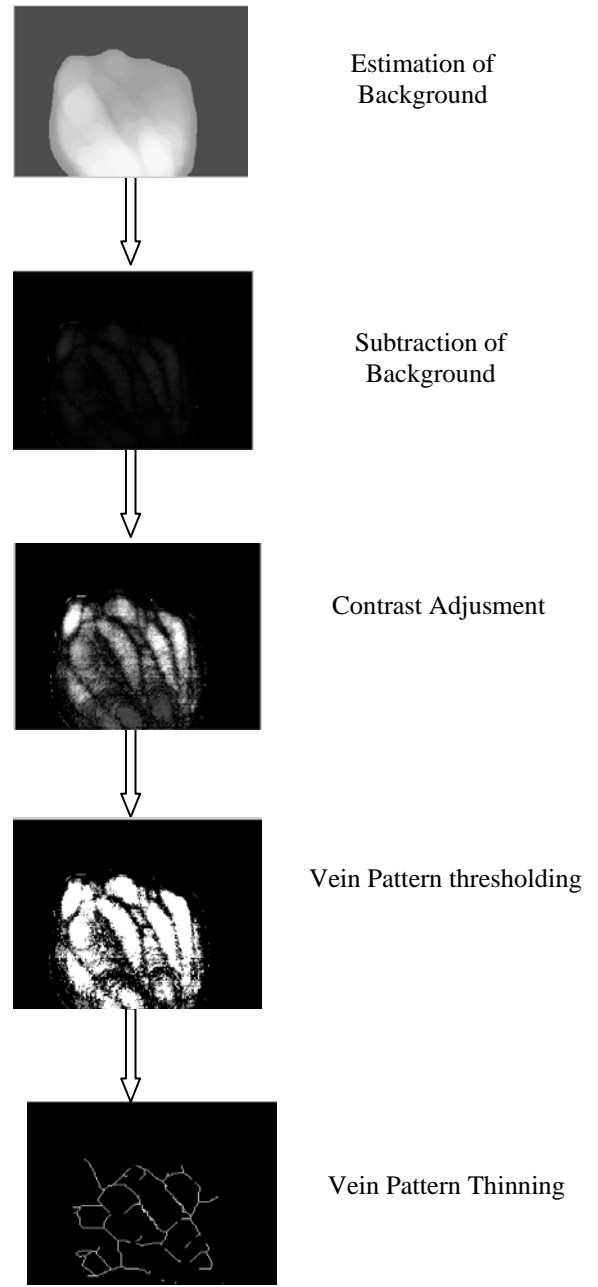

Figure 2. Biometric Procedure

### C. Enhancement and Thinning of the Vein Pattern

The clearness of the vein pattern varies from image to image. Thus, we had to enhance the quality of the image to obtain the vein structures. To achieve this, different filters was applied on these vein patterns segmented. Match filter, Wiener filter and smoothing filter as proposed by S.Zhao et al, [14] was used to suppress noises that exist in the vein pattern. This allowed us to obtain clearer vein pattern for feature extraction.

As the size of veins grow as human beings grow, only the shape of the vein pattern is used as the sole feature to recognize each individual. A good representation of the pattern's shape is via extracting its skeleton. A thinning algorithm was devised to obtain a thinned version of the vein pattern. Pruning, which eliminates the shadow in the images was applied on the image.

III. FEATURE EXTRACTION OF THE VEIN PATTERN USING QUADRATIC INFERENCE FUNCTION (QIF)

After obtaining the vein pattern, the coordinates were extracted from the pattern. Each coordinate represent the pixel values of the image

We assume I images for the training set X, i.e,

$$X = [X_1, X_2, \ldots X_i, \ldots X_I] \quad (2)$$

$$X_i = \begin{bmatrix} X_{i11} & X_{i12} & \ldots & X_{i1N} \\ X_{i21} & X_{i22} & \ldots & X_{i2N} \\ \vdots & \vdots & X_{ijk} & \vdots \\ X_{iM1} & \ldots & \ldots & X_{iMN} \end{bmatrix} \quad (3)$$

where

$$X_{ijk} = (x_{ij}, y_{ik}) \quad (4)$$

where $i$ is the index for the $i^{th}$ image, $j$ is the corresponding index for the $x$-coordinate of the $i^{th}$ image, $k$ is the corresponding index for the $y$-coordinate of the $i^{th}$ image where $i = 1, \cdots, I$, $j = 1, \cdots, M$ and $k = 1, \ldots, N$. Thus, the training matrix $X$ is of dimension $M \times 2NI$. Note that for each image $X_i$ in the training set $X$, the number of $x$ co-ordinates is chosen using the condition $M = \min(M_1, M_2, \ldots, M_i, \ldots, M_I)$, where $M_i =$ the number of $x$ co-ordinates in image $i$ and the number of $y$ co-ordinates is chosen using the condition $N = \min(N_1, N_2, \ldots, N_i, \ldots, N_I)$, where $N_i =$ the number of $y$ co-ordinates in image $i$. These two conditions are very important in stabilizing the dimensionality of the matrix $X_i$.

However, this way of representing the training set has two major drawbacks. Firstly, there may be variation in the elements of the training set and secondly, the dimensions of the matrix is very large especially when $I$ is large. Thus, it becomes difficult to work with the original training matrix $X$. To solve these problems, we standardize the coordinates by averaging, i.e,

$$\psi_j = \frac{1}{I} \sum_{i=I}^{I} x_{ij} \quad (5)$$

$$\text{and } \psi_k = \frac{1}{I} \sum_{i=I}^{I} y_{ik} \quad (6)$$

where $\psi_{jk} = (\psi_j, \psi_k)$.

Let $\phi_{ij} = x_{ij} - \psi_j$ and $\phi_{ik} = y_{ij} - \psi_k$. Thus, we can write $\phi_{ijk} = X_{ijk} - \psi_{jk}$, we then construct another training set $\phi = [\phi_1, \phi_2, \ldots, \phi_i, \ldots, \phi_I]$ in the same way as the training set $X$ using the new co-ordinate systems defined in equations (5) and (6). To reduce the dimension and avoid the original training matrix $X$, we construct a new matrix of lower dimension that is based on the following quadratic inference formula [8],[17]:

$$Q = g^T C^{-1} g \quad (7)$$

where g is a matrix of dimension $M \times 2N$ and its elements are $g_{jk} = \psi_{jk}$ and $C = \frac{1}{I} \sum_{i=1}^{I} \phi_i \phi_i^T$.

Note the dimension of the training set has been from $M \times 2NI$ to $2N \times 2N$ following equation (7). To generate the eigenvein, we use

$$Qv_i = \mu_i v_i \quad (8)$$

For each eigenvector, a family of eigenvein has to be generated. However, many eigenveins are being generated. In order to determine how many eigenveins are required, the following formulae are being used. We have accounted for 90 % and 95% of the variation in the training set.

$$\frac{\sum_{i=1}^{2N'} \mu_i}{\sum_{j=1}^{2N} \mu_j} > 0.9 \quad (9)$$

$$\frac{\sum_{i=1}^{2N'} \mu_i}{\sum_{j=1}^{2N} \mu_j} > 0.95 \quad (10)$$

We have already obtained $2N'$ eigenveins. For each element in the training set, the weight is calculated. This weight will demonstrate the contribution of each eigenvein to respective training element. If the weight is bigger, then the eigenvein has shown the real vein. If the value is less, there is no big contribution with the real vein for that particular eigenvalue. The following operation shows how each element in the training set is projected onto the vein space:

$$\omega_k = \frac{1}{NM} \sum_{i=1}^{N} \sum_{j=1}^{M} (Q\upsilon_k)^T (X_{ij}^T - \phi_j^T) \quad (11)$$

where,

$$1 < k < 2N', 1 \leq i \leq 2N, j = 1...M$$

Each element in the training set has a weight to determine their contribution to the vein space.

### IV. VEIN PATTERN MATCHING

To recognize an image means to check whether the image exist in the database. When a person wants to get access to the system, the picture of the vein, known as the test image is captured. The coordinates of the test image are obtained and represented as the training set. The weight of the new image is calculated and projected on the vein space [15],[16]. The vein space contains all the vein images. Thus, we have to check whether the input image exist in that space. The Euclidean distance between the projected image and those stored is being calculated. First of all, our system checks whether the test image is a vein by testing it with an arbitrary value. Then the Euclidean distance is computed to check whether the test image exist in the database. If it is vein image, then it is accepted. The results were recorded and analyzed.

### V. EXPERIMENTAL RESULTS

The hand dorsal vein biometric was tested using pixel by pixel method and the quadratic inference method discussed in this paper. It is to be noted that pixel by pixel method test each individual pixel by counting the number of overlapped pixel in the test image and that of the template found in the database.

In order to test the efficiency and accuracy of the method proposed, false acceptance rate (FAR) and false rejection rate(FRR) are computed. False Acceptance Rate refers to the total number of unauthorized persons getting access to the system over the total number of people attempting to use the system. False Rejection Rate refers to the total number of authorized persons not getting access to the system over the total number of people attempting to get access to the system. The table below shows the FAR and FRR for 20,40,60,80 and 100 images tested.

TABLE I. FAR AND FRR USING PIXEL BY PIXEL METHOD

| Number of images | FAR(%) | FRR(%) |
|---|---|---|
| 20 | 0.1000 | 0.1500 |
| 40 | 0.0250 | 0.0750 |
| 60 | 0.0340 | 0.0670 |
| 80 | 0.0375 | 0.0250 |
| 100 | 0.0400 | 0.0300 |

TABLE II. FAR AND FRR USING QUADRATIC INFERENCE FUNCTION

| Number of images | FAR(%) | FRR(%) |
|---|---|---|
| 20 | 0.0500 | 0.0600 |
| 40 | 0.0250 | 0.0500 |
| 60 | 0.0340 | 0.0340 |
| 80 | 0.0250 | 0.0125 |
| 100 | 0.0200 | 0.0300 |

According to the results obtained, the FAR and FRR is less when using quadratic inference function compared to pixel by pixel method. In order to test the efficiency of our proposed method, we have computed the matching time of the method illustrated in the table below:

TABLE III. COMPARISON OF MATCHING TIME

| Number of images | Matching time using pixel by pixel(in second) | Matching time using Quadratic Inference function (in second) |
|---|---|---|
| 20 | 275 | 130 |
| 40 | 580 | 300 |
| 60 | 843 | 450 |
| 80 | 1130 | 575 |
| 100 | 1400 | 700 |

From the results obtained, it is noticed that the matching time of the proposed method is less compared to the pixel by pixel method. According to the experimental results, quadratic inference function method is on average twice faster compared to pixel by pixel method.

### VI. CONCLUSION

The new method proposed that is the quadratic inference function was successfully applied on hand dorsal vein pattern providing satisfactory results. The FRR and FAR were computed and are found to be less when using our proposed method. It also reduces the dimension of the matrices which consequently has an impact on matching time. The matching time is improved in our proposed method and this is desired in all biometric security system.

### ACKNOWLEDGMENT

We express our deepest thanks to Prof. Ahmed M. Badawi, from University of Tennessee, Knoxville, for providing us with a dataset of 200 images of hand dorsal vein pattern.

.